\documentclass{article}
\usepackage[preprint]{neurips_2020}

\usepackage[utf8]{inputenc} %
\usepackage[T1]{fontenc}    %
\usepackage{booktabs}       %
\usepackage{amsfonts}       %
\usepackage{nicefrac}       %
\usepackage{amsmath}
\usepackage{amssymb}
\usepackage{amsthm}
\usepackage{bm}
\usepackage{enumitem}
\usepackage{multirow}
\usepackage{graphicx}
\usepackage{subcaption}
\usepackage{makecell}
\usepackage{array}
\usepackage{url}
\usepackage{algorithm}
\usepackage{algorithmic}
\usepackage{dsfont}
\usepackage{bbm}
\newtheorem{theorem}{Theorem}
\newtheorem{proposition}{Proposition}
\newtheorem{corollary}{Corollary}

\usepackage{listings}
\usepackage{commands}
\usepackage{xcolor} 
\usepackage{xspace} 

\title{Robust Flow-based Conformal Inference (FCI) with Statistical Guarantee}

\author{%
  Youhui Ye\\
  Department of Statistics\\
  Virginia Tech, USA\\
  \texttt{yye1997@vt.edu}
  \And
  Meimei Liu \\
  Department of Statistics \\
  Virginia Tech, USA \\
  \texttt{meimeiliu@vt.edu} \\
  \AND
  Xin Xing \\
  Department of Statistics \\
  Virginia Tech, USA \\
  \texttt{xinxing@vt.edu} \\
}

\begin{document}

\maketitle

\begin{abstract}
Conformal prediction aims to determine precise levels of confidence in predictions for new objects using past experience. However, the commonly used exchangeable assumptions between the training data and testing data limit its usage in dealing with contaminated testing sets. In this paper, we develop a novel flow-based  conformal inference (\FCI) method to  build predictive sets and infer  outliers for complex and high-dimensional data. We leverage ideas from adversarial flow to transfer the input data to a random vector with known distributions. Our roundtrip transformation can map the input data to a low-dimensional space,  meanwhile reserve the conditional distribution of input data given each class label, which enables us to construct a non-conformity score for uncertainty quantification.
Our approach is applicable and robust when the testing data is contaminated. We evaluate our method, robust flow-based conformal inference, on benchmark datasets. We find that it produces effective predictive sets and accurate outlier detection and is more powerful relative to competing approaches. 
\end{abstract}

\section{Introduction}
Deep learning algorithms are increasingly being applied in modern decision-making processes. For example, self-driving cars detect objects using complex data from cameras and other sensors \citep{rao2018deep}. Intelligent disease diagnosis use data from various types of medical imaging to clinical records \citep{bakator2018deep}. In such cases, the test data usually contains noisy or contaminated observations or outliers that have not been seen in limited training data. It is crucial to develop robust methods to quantify the uncertainty of their predictions in the decision-making process.

Conformal prediction constructs valid prediction intervals instead of point estimates for new objects using past data. It was introduced by \cite{vovk2005,vovk2012conditional}, and has a wide application in regression \citep{lei2014distribution}, classification \citep{Romano2020}. 
The classical conformal prediction uses scores 
outputted by black-box classifiers such as deep neural networks to build the confidence sets. However, the 
probabilities output by convolutional neural networks (CNNs) is overconfidence or underconfidence, which is observed as the probability values concentrated to 0 or 1 \citep{nixon2019measuring}, leading to inaccurate conformal prediction. In order to calibrate the probabilities, data-splitting is one commonly used strategy to estimate the distribution of input data \citep{Romano2020,angelopoulos2020uncertainty}. It uses the first half of the data to train the model and uses the second half of the data to evaluate the distribution and build prediction set, which inevitably causes power loss. Data-augmentation conformal inference is Another popular approach to construct  prediction set is via data-augmentation conformal inference. These approaches can guarantee the coverage level is controlled provided that the training and test data satisfies  exchangeable condition. 

In this paper, we develop a flow-based conformal inference (\FCI) to construct prediction intervals for target data with  statistical guarantee. 
A notable feature of \FCI is its ability in learning the distribution of the complex training data for each class. To achieve this feature, we focus on learning a roundtrip transformation between the input data (viewed as the real domain) and its latent domain via the adversarial generative model \citep{Goodfellow2014}. By adding the cycle consistency constraint \citep{Zhu2017}, the transformation can reserve the conditional distribution of our complex input data given each class label. %
Then, to develop the predictive region, we consider testing whether a prediction is from a given class. Leveraging on the learned latent domain, we construct test statistics with an asymptotic Chi-squared distribution and derive a valid $p$-value that enjoys the controlled type-I error rate. We also show that the proposed $p$-value can serve as a basis for various inference tasks, including building predictive sets and outlier detection.

We further consider the cases where the outliers are presented in the test data. For example, in diagnosis, the patient in the testing set may have a rare disease that deviates from the training data. It is harmful to give a wrong predictive set, including other types of disease or health. Instead, detecting it as an outlier for the doctor's further review is a better decision. However, classical conformal predictions are prone to be unstable in such cases since the commonly used assumption that the training and test data are exchangeable  \citep{gibbs2021adaptive} is not valid when presenting outliers in test data. We remark that the proposed FCI does not rely on the exchangeable assumption, thus is feasible for outlier detection.  
We can formulate the outlier detection through the corresponding test statistics characterization. Depending on the distinguishability of the test statistics, our proposed test is sensitive to detect the outliers that deviation from the null distribution.  %

{\bf{Related Work}} In classical statistical approaches, Bayesian neural networks  \citep{neal2012bayesian} and bootstrap methods \citep{gupta2022nested} are often used to quantify uncertainty. These approaches do not scale well on large datasets due to the computational bottleneck in the sampling process. Recently, conformal prediction has been an approach for generating predictive sets that satisfy the coverage property. The most popular approach in this line is a data-splitting version known as the split conformal prediction that enables conformal prediction methods to be deployed for any classifier \citep{papadopoulos2002inductive,lei2014distribution,angelopoulos2020uncertainty}. This line of work assumes the test data and training data are exchangeable, which limits its application when there are outliers or distributional shifts in the test data. \citet{gibbs2021adaptive} introduces a tunable parameter for adaptive data-splitting methods to deal with the distributional change. However, it is still challenging to handle cases where outliers appear.

Probability distribution is an essential tool to quantify the uncertainty in various modern inference tasks. Compared with classical statistical methods such as Bayesian inference \citep{meng1994posterior}, nonparametric density estimation \citep{davis2011remarks} and bootstrap \citep{stine1985bootstrap}, flow-based generative models leverage the ability to learn the distribution in complex data in a scalable fashion. Normalizing flows represent learning an invertible transformation from input data to a latent variable with a known density. To achieve this, the neural density estimators have to impose heavy constraints. 
To mitigate this issue, \citet{liu2021density} employ the cycle-consistency constraint in cycleGAN \citep{Zhu2017} to learn a pair of transformations between the latent variable space and the data space, which demonstrates state-of-the-art performance in various applications. Motivated by the power of the flow-based method in learning complex distribution, we propose a conditional adversarial flow network as a basis for our conformal inference framework.

\section{Method}

Denote the training data as $\{(X_i,Y_i)\}_{i=1}^n$, where $Y_i\in \{1,\cdots, L\}$ is the outcome of interest and $X_i\in \mathcal{X}$ is the predictor. Given the training data, we aim to quantify the uncertainty on a new i.i.d. input $\Xnew$ with a prediction set of possible labels $\widehat{C}_n : \mathcal{X} \to \{\textrm{subset of } \{1,\cdots,L\}\}$ satisfying 
$$
P\{ \Ynew \in \widehat{C}_n(\Xnew)\} \geq 1-\alpha,
$$
where $1-\alpha \in (0,1)$ is a desired coverage rate. Constructing $\widehat{C}_n$ is related to a fundamental statistical problem for testing whether an observation $\Xnew$ belongs to the $\ell$th class for $\ell \in \{1,\dots, L\}$. Denote $F_\ell$ as the distribution of $\Xnew$ given $\Ynew = \ell$. We consider the following hypothesis 
\begin{equation}\label{eq:hypothesis}
    H_0: \Xnew \sim F_\ell \quad\mbox{ v.s. }\quad H_1: \Xnew \nsim F_\ell.
\end{equation}
The hypothesis testing in (\ref{eq:hypothesis}) paves the way for various inference tasks, including building conformal predictive sets and detecting outliers. One way to construct the conformal set is via $p$-value under $H_0$ is true.

Denote $\alpha\in(0,1)$ as the predefined type-I error which is also known as the false positive rate and $\pi_\ell(\Xnew)$ as the the $p$-value for hypothesis testing (\ref{eq:hypothesis}) satisfying 
\begin{equation}\label{eq:typei}
P(\pi_\ell(\Xnew) \leq \alpha |Y_i = \ell) \leq \alpha 
\end{equation}
i.e., the $p$-value is uniformly distributed in $(0,1)$ if $H_0$ is true. The $p$-value characterizes the probability of obtaining a test result at least as extreme as the result observed, which can be viewed as the score for conformal learning. With $p$-values known, we can construct the predictive set for $\Xnew$ as 
\begin{equation}\label{eq:pvalue_set}
\widehat{C}_n(\Xnew) = \{\ell: \pi_\ell(\Xnew)\geq \alpha\}. 
\end{equation}
For example, suppose $X_i$'s are one-dimensional random variables following known Gaussian distribution with mean $\mu_\ell$ and variance $\sigma_\ell^2$  conditioned on $Y_i=\ell$. We then have the explicit form of $p$-value for $\Xnew$, i.e., 
$\pi_\ell(\Xnew) = 2\min\{\Phi((\Xnew - \mu_{\ell}) / \sigma_{\ell}), 1-\Phi((\Xnew - \mu_{\ell}) / \sigma_{\ell})\}$ where $\Phi$ is the CDF of standard normal distribution. In this case, the univariate Gaussian distribution is unimodal with a light tail, which makes the $p$-value decrease exponentially for a value far from the mean and thus tends to have a high power to detect the extreme values. The predictive set for $\Xnew$ follows (\ref{eq:pvalue_set}). In practice, however, we do not have direct access to $F_\ell$ for high-dimensional or complex covariates $X$, leading to a challenge problem to find a function of $\{X_i\}_{i=1}^n$ as a valid $p$-value satisfying (\ref{eq:typei}).

Instead of estimating the density function of $X$ directly, we aim to evaluate the hypothesis testing (\ref{eq:hypothesis}) through a lower-dimensional latent space, facilitating the construction of conformal sets. Traditional dimensional reduction methods often compress the data without retaining the density information. Fortunately, the recent development of adversarial generative models \citep{Zhu2017} enables the power to learn the distribution of complex data via a transformer from its low-dimensional embedding. Without loss of generality, denote $X\in \cX$ as the predictor, and $\bZ\in \cZ\subset \mathbb{R}^d$ as a random variable with known distribution. Based on GAN architecture, we design a roundtrip between $X$ and $\bZ$ via two functions $G(\cdot): \cZ \to \cX$ and $I(\cdot): \cX \to \cZ$ with $G(\bZ)=X$ and $I(X)=\bZ$. Suppose $d< \textrm{dim}(X)$, $\bZ$ can be viewed as the latent representation of $X$, and such transformation can reserve the conditional distribution of $X$ given each class label. 
Therefore, %
leveraging on $\bZ$, we develop a novel $p$-value for hypothesis problem in (\ref{eq:hypothesis}) which serves as a basis for our conformal inference framework.

In the following Section \ref{subsec:caf}, we show the designed bidirectional conditional GAN architecture; in Section \ref{subsec:pvalue}, we construct valid $p$-values as the non-conformity score based on the latent representations of $X$ and build its asymptotic property.

\subsection{Conditional Adversarial Flow}\label{subsec:caf}

\begin{figure*}[ht]
\begin{center}
\includegraphics[width = 0.85\textwidth]{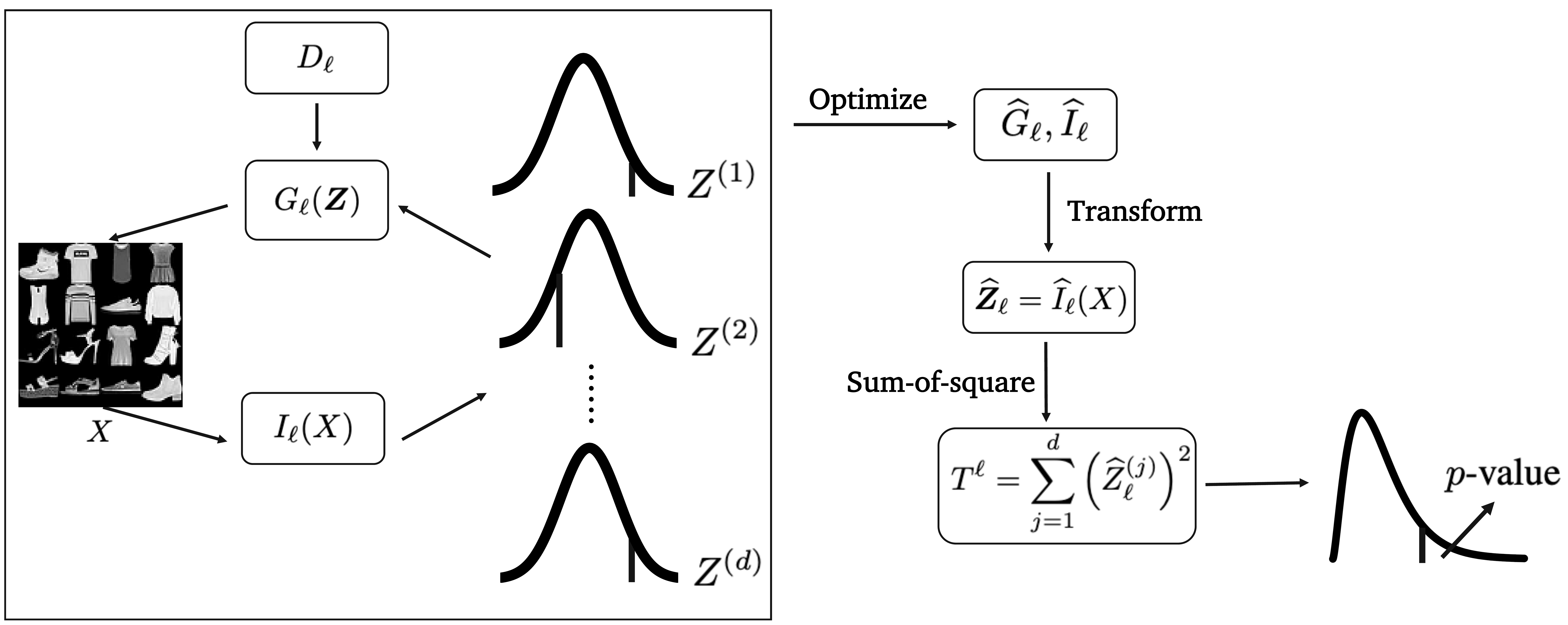}
 \caption{\footnotesize Illustration of flow-based conformal inference for image data. $\bZ = (Z^{(1)}, Z^{(2)}, \dots, Z^{(d)})$ follows multivariate Gaussian distribution with mean zero and identity covariance matrix. $T^\ell$ is our proposed non-conformity score constructed by latent representation $\widehat{\bZ}_\ell=\widehat{I}_\ell(X)$. We construct one-sided $p$-value for hypothesis testing problem in (\ref{eq:hypothesis}) based on the distribution of $T^\ell$.} 
\label{fig:GAN} 
\end{center}
\end{figure*}

We first develop a conditional adversarial flow to characterize the conditional distribution of $X$ given $Y=\ell$ via a roundtrip mapping $G_\ell: \cZ \rightarrow \cX$ and $I_\ell : \cX \rightarrow \cZ$.  \citet{Zhu2017} introduced cycle consistent adversarial networks for unpaired image translation tasks. We employ the cycle consistency loss to transfer the input data $X$ to $\bZ$. Particularly, our roundtrip model consists of two generative adversarial network (GAN) models: (1) a forward GAN model $G_\ell$ aims at generating samples $G_\ell(\bZ)$ that are similar to observation data $X$ equipped with a discriminator $D_\ell$ to distinguish observation data from generated samples; (2) a backward GAN model $I_\ell(X)$ to transform the data distribution to approximate the base distribution in latent space; see the left part of Figure \ref{fig:GAN} for illustration. 

In this paper, we focus on the image data using CNN structures; however, the framework can be generalized to model other data types by adjusting the neural network architecture.

{\bf Learning the forward generative model $(G_\ell(\cdot), D_\ell(\cdot))$ } 
We propose to construct our generator based on a series of transposed convolutional filters \citep{dumoulin2016guide}. Consider $\bZ$ from a standard multivariate Gaussian distribution with density function $P_{\bZ}(\bz)$ defined as 
\begin{equation*}
P_{\bZ}(\bz)  = (\sqrt{2\pi} )^{-d} \exp\left(-\frac{||\bz||_2^2}{2}\right), 
\end{equation*}
where $||\cdot||_2$ is the $L^2$ norm, and $d$ is the dimension of $\bZ$. The generator $G_\ell$ can be represented as 
$$
G_\ell = C^{(K)} \circ \cdots \circ C^{(1)}(\bZ),
$$
where $C^{(k)} = (C^{(k)}_1,\dots, C^{(k)}_{q_k})$ denotes the transposed convolutional layer with $q_k$ channels. For example, the first transposed convolutional layer $C^{(1)}(\bZ)$ is based on each element of the vector $\bZ$. Then the output of the first transposed convolutional layer  is 
\begin{equation*}
\begin{split}
    C^{(1)} & = (C^{(1)}_1, C^{(1)}_2, \dots, C^{(1)}_{q_1}), \\
    \text{with} \quad C^{(1)}_t & = \sigma(\omega_{1t} \odot \bZ) \quad \mbox{for } t=1,\dots, q_1,
\end{split}
\end{equation*}
where $\omega_{1t}$ is the $t$th filter in layer 1, and $\sigma(\cdot)$ is the activation function. To achieve the size of the target image $X$, we sequentially apply the transposed convolutional filters
\begin{equation*}
\begin{split}
    C^{(k+1)} & = (C^{(k+1)}_1, C^{(k+1)}_2, \dots, C^{(k+1)}_{q_{k+1}}), \\
    \text{with} \quad C^{(k+1)}_t & = \sigma(\omega_{k+1, t} \odot C^{(k)}) \quad \mbox{for } t=1,\dots, q_{k+1},
\end{split}
\end{equation*}
until the output dimension is equal to the image dimensions. We denote the output as $C^{(K)}$. The discriminator
network $D_{\ell}: \cX \to \{0,1\}$ aims to distinguish whether the generated image is real or fake, and can be viewed as a binary classifier with the output as the probability that the data point comes from the real observation rather than the generated sample.

We express the objective as
\begin{multline}\label{eq:valueG}
    V_{G}(G_\ell,D_{\ell})  =  \bbE_{X\sim F_\ell}[\log D_{\ell}(X)] \\
 + \bbE_{\bZ\sim P_{\bZ}}[\log(1- D_{\ell}(G_\ell(\bZ)))],
\end{multline}
where the generator $G_\ell$ aims to maximize the failure rate of the discriminator, while the discriminator $D_{\ell}$ aims to minimize itself. As shown in \citet{Goodfellow2014}, the two networks are locked in a two-player minimax game defined by the value function $V_{G}$.

{\bf Learning the backward generative model  $I_\ell(\cdot)$} 
We consider the base distribution of $\bZ$ as multivariate Gaussian with an identity covariance matrix since we aim to minimize the dependency among the entries in $\bZ$. 

Instead of using a discriminator in the backward GAN model, we use a distance measure $\cM$ for distributions to guarantee the distribution of transformed vector  $I_\ell(X)$ to be close to the base distribution of $Z$. 
The maximum mean discrepancy (MMD) \citep{gretton2012kernel} is a popular and suitable tool. Suppose there are two probability measures $\mu$ and $\nu$ on the latent space $\cZ$. Given $U$ and $U'$ independent variables with probability measure $\mu$, and $V$ and $V'$ independent variables with probability measure $\nu$, the squared MMD is
\begin{equation*}
\begin{split}
    \cM_k^2 [\mu, \nu] & = \EE_{U, U'}[k(U, U')] \\
     & - 2\EE_{U, V}[k(U, V)] + \EE_{V, V'}[k(V, V')]
\end{split}
\end{equation*}
where $k(\cdot, \cdot): \cZ \times \cZ \rightarrow \RR$ is a kernel function. To ensure that $\mu=\nu$ is equivalent to $\cM_k^2[\mu, \nu]=0$, we require that the kernel is a characteristic kernel, i.e., the mean embedding of $k(\cdot, \cdot)$, $\mu \to \EE_{U\sim \mu}[k(U, \cdot)]$, is an injective mapping \citep{gretton2012kernel}. One example of characteristic kernel is Gaussian kernel $k(u, v) = \exp(||u-v||^2)$. \cite{gretton2012kernel} also provides an unbiased empirical estimate for observations $U=(U_1,\dots, U_m)$ and $V=(V_1, \dots, V_n)$ as
\begin{multline*}
        \widehat{\cM}^2_k [U, V] =  \frac{1}{m(m-1)} \sum_{i=1}^m \sum_{j\neq i}^m k(U_i, U_j) \\
        + \frac{1}{n(n-1)} \sum_{i=1}^n \sum_{j\neq i}^n k(V_i, V_j)  - \frac{2}{mn} \sum_{i=1}^m \sum_{j = 1}^n k(U_i, V_j).
\end{multline*}
Note that $\widehat{\cM}_k(U, V )$ may not be zero even when $\mu=\nu$ due to the sampling variance. Recent theoretical studies  \citep{simon2020metrizing, simon2018kernel} show that MMD-distance metrizes the weak convergence of probability measures which enables us to derive the asymptotic distribution of our proposed non-conformity score later. Therefore, we 
set the objective to learn $I_\ell$ through the MMD value function:
\begin{equation}\label{eq:vi}
V_I(I_\ell) = \cM_k^2[P_{I_\ell(X)}, P_{\bZ}] .
\end{equation}
where the inverse generator $I_\ell$ aims to make the distribution of $I_\ell(X)$ close to $P_{\bZ}$.
Practically, we replace $\cM_k^2$ with $\widehat{\cM}_k^2$ using the observations $X_1, X_2, \dots, X_n$ for $I_\ell(X)$ and the sample drawn from $P_{\bZ}$. 

{\bf Cycle consistency loss}
We aim to minimize the roundtrip loss, that is, the distance when a data point goes through a roundtrip transformation between two data domains. As demonstrated in \citet{Zhu2017}, it is necessary to supplement a cycle consistency loss, to guarantee 
$X \to I_\ell(X) \to G_\ell(I_\ell(X))$ be close to $X$ , and $\bZ \to G_\ell(\bZ) \to I_\ell(G_\ell(\bZ))$ be close to $\bZ$. We formulate the cycle consistency loss function as
\begin{align}
    V_{\rm cycle}(G_\ell, I_\ell)  = &  \bbE_{X\sim F_\ell}[\|X - G_\ell(I_\ell(X)) \|_2]\nonumber\\
&  + \bbE_{\bZ\sim P_{\bZ}}[\|\bZ - I_\ell(G_\ell(\bZ)) \|_2]\label{eq:valuecycle} .
\end{align}

{\bf{Total training loss}} 
To train the three networks $G_\ell$, $I_\ell$, and $D_{\ell}$, we combine the adversarial training loss  (equation (\ref{eq:valueG}) and equation (\ref{eq:vi}))  and cycle consistency loss (equation (\ref{eq:valuecycle})) together, and get the total training loss as 
\begin{equation*}
\begin{split}
    V_{\rm total}(G_\ell, D_{\ell}, I_\ell) = & V_{G}(G_\ell,D_{\ell}) + V_I(I_\ell) \\
    & + V_{\rm cycle}(G_\ell, I_\ell). 
\end{split}
\end{equation*}
The estimates $\widehat{G}_\ell$ and $\widehat{I}_\ell$ can be achieved by optimizing $V_{\rm total}$. That is,    
\begin{equation}\label{eq:estimation}
    \widehat{G}_\ell, \widehat{I}_\ell = \arg \min_{G_\ell, I_\ell}\max_{D_{\ell}} V_{\rm total}(G_\ell, D_{\ell}, I_\ell). 
\end{equation}

In the following Section \ref{subsec:pvalue} and \ref{subsec:pre set}, we construct the non-conformity score and predictive set for $X_{\textrm{new}}$ leveraging on the learned $\widehat{G}_\ell$ and $\widehat{I}_\ell$.

\subsection{Construct $p$-value and asymptotic property}\label{subsec:pvalue}

{\bf Construct the non-conformity score} 
Denote the transformed vector through $\widehat{I}_\ell$ as %
$\widehat \bZ_{i, \ell} = \widehat{I}_\ell(X_i)$ for $\ell=1,\dots, L$.
We apply the {\it sum-of-square} function to $\widehat{\bZ}_{i, \ell}$ inspired by the Chi-squared test statistic to construct the non-conformity score. As introduced in the second and third chapters \citep{vovk2005}, a non-conformity score measures how dissimilar one example is from a set of examples. 
Let $\widehat \bZ_{i, \ell} = (\widehat Z_{i, \ell}^{(1)}, \widehat Z_{i, \ell}^{(2)}, \dots, \widehat Z_{i, \ell}^{(d)})^\top$, we define the non-conformity score as
\begin{equation*}\label{eq:Tscore}
    T^\ell(X_i) = \sum_{j=1}^d \left(\widehat Z_{i, \ell}^{(j)}\right)^2 .
\end{equation*}
For a new observation $\Xnew$, we call $\Tnew^\ell = T^\ell(\Xnew)$ as its non-conformity score for the $\ell$th class. Then for each $\Xnew$, there are $L$ non-comformity scores in total: $\Tnew^1, \Tnew^2, \dots, \Tnew^L$. 
Take $\ell = 1$ as an example, a higher non-conformity score $\Tnew^{1}$ implies a larger difference between the new observation $\Xnew$ and the set $\{ X_i: Y_i = 1 \}$; then we tend to exclude the class $\ell$ from its prediction set in conformal inference.  

In order to derive the asymptotic distribution of $\Tnew^\ell$ under null hypothesis $H_0$ defined in (\ref{eq:hypothesis}), we consider the case where the total loss converges, i.e., $V_{\rm total } \to 0$ as $n\to \infty$. As shown in \cite{gretton2012kernel}, MMD can separate all probability
distributions $\mu$, $\nu$.  $\cM_k^2(\mu, \nu) = 0$ if and only if $\mu = \nu$ for any characteristic kernel $k(\cdot, \cdot)$. A natural question is whether $P_{\widehat{\bZ}}$ converges to $P_{\bZ}$ under the convergence of the total loss, which is at the heart of characterizing the asymptotic distribution of $T^\ell(X_{\rm new})$.  We introduce the result in \cite{simon2018kernel} which characterizes the metrization of weak convergence of probability measures or convergence in distribution.

\begin{proposition}\label{prop:mmd}
(\cite{simon2018kernel}) On a locally  compact Hausdorff space, a bounded, Borel measurable kernel metrizes the weak convergence of probability measures if and only if it is continuous and characteristic (to the set of probability measures).
\end{proposition}

As a consequence of the Heine–Borel theorem, the space of $\bZ$ is a locally compact Hausdorff space. 
Proposition \ref{prop:mmd} suggests us to choose a continuous and characteristic kernel function $k(\cdot,\cdot)$ in learning $I_\ell$ to guarantee $P_{\widehat{\bZ}}$ converges to $P_{\bZ}$. Next, we show the asymptotic distribution of $T^{\ell}(X_\textrm{new})$ under $H_0$.

\begin{theorem}\label{thm:asymptotic}
({\bf Asymptotic distribution})
Assume that $V_{\rm total} \to 0$ as $n\to \infty$ and $k(\cdot, \cdot)$ is continuous and characteristic. Under $H_0: \Xnew\sim F_{\ell}$, we have 
\begin{equation*}
    T^\ell(\Xnew) \xrightarrow{D} \chi^2_d, \quad n \to \infty, \ell = 1,\dots, L.
\end{equation*}
where $\xrightarrow{D}$ denotes the convergence in distribution.
\end{theorem}

The proof is included in the Supplementary. Theorem \ref{thm:asymptotic} shows that our proposed non-conformity score converges to a Chi-squared distribution which is a unimodal probability distribution that guarantees the extreme value is on the tail. Based on the unimodality of the asymptotic distribution,  we construct the one-side $p$-value as 
\begin{equation}\label{eq:pi}
    \pi^\ell_{\rm new} = \pi^\ell(\Xnew) = \frac{ \left|\{T_i^\ell \in A_\ell: \Tnew^\ell \geq T_i^\ell\} \right|}{|A_\ell|},
\end{equation}
where 
\begin{equation}\label{eq:pool}
    A_\ell = \{ T_i^\ell = T^\ell(X_i): Y_i = \ell\}, 
\end{equation}
and $|\cdot|$ denotes the cardinality operation.

\begin{corollary}\label{col:pvalue}
 Assume that $V_{\rm total} \to 0$ as $n\to \infty$ and $k(\cdot, \cdot)$ is continuous and characteristic. We have
\begin{equation*}
P(\pi_{\rm new}^\ell \leq \alpha |Y_{\rm new} = \ell) \leq \alpha.
\end{equation*}
as $n\to\infty$.
\end{corollary}
Corollary \ref{col:pvalue} shows that our proposed $p$-value can control the type-I error of the hypothesis in (\ref{eq:hypothesis}). This Corollary forms the basis for constructing the predictive set in Section \ref{subsec:pre set}.

\subsection{Construct the predictive set}\label{subsec:pre set}

To characterize the uncertainty of prediction for $\Xnew$, our goal is to construct a predictive set $\widehat{C}_n(X_{\rm new})$ such that $\Ynew$ belongs to $\widehat{C}_n(X_{\rm new})$ with probability large than or equal to $1-\alpha$ where $\alpha\in (0,1)$ is a predefined coverage level. 

We use the $p$-value defined in (\ref{eq:pi}) as our uncertainty quantification  measure; that is, a smaller $p$-value indicates stronger evidence to reject the hypothesis in (\ref{eq:hypothesis}).
Then, we can construct a prediction set
\begin{equation}\label{eq:predset}
\widehat{C}_n(\Xnew) = \{\ell:\pi^\ell_{\rm new} \geq \alpha \}, 
\end{equation}
 for any predefined coverage level $\alpha\in(0,1)$. We call $\widehat{C}_n(\Xnew)$ as the flow-based conformal inference (FCI). For prediction task, it is usually hard to show the convergence of accuracy to the best of our knowledge. However, there are a lot of works \citep{vovk2005, papadopoulos2008} on the coverage probability in literature. We mainly focus on this task, and show the convergence of the coverage probability as follows.

\begin{theorem}\label{thm:coverage}
(\FCI predictive set coverage guarantee). If $\Ynew\in \{1,\dots,L\}$, we assume that that $V_{\rm total} \to 0$ as $n\to \infty$ and $k(\cdot, \cdot)$ is continuous and characteristic. Then for the predictive set defined in (\ref{eq:predset}), 
\begin{equation*}
    P(\Ynew \in \widehat{C}_n(\Xnew)) \geq 1-\alpha
\end{equation*}
as $n\to \infty$.
\end{theorem}
Theorem 2 shows that \FCI can achieve pre-specified coverage probability.

\subsection{Robustness under the contamination} 
In order to show the robustness of the proposed method, we consider a common scenario where the test dataset is contaminated, i.e., existing data points that differ significantly from observations in training. In such a scenario, the exchangeable assumption between the training and test is invalid.
The flow-based model described above can be adapted to this scenario since we learned all the conditional distributions in each class in the training. If all the hypothesis in (\ref{eq:hypothesis}) for $\ell = 1,\dots, L$ has been rejected, it means the point is extreme for all the training classes. Then we conclude that the new data point is an outlier and use our $p$-values to control the type-I error rate, i.e., the rate of false detection of outliers. 

\begin{corollary}\label{col:outlier}
(Type-I error of \FCI outlier detection). Assume that $V_{\rm total} \to 0$ as $n\to \infty$ and $k(\cdot, \cdot)$ is continuous and characteristic. If $\Ynew \in \{ 1,\dots, L\}$, then we have  (\ref{eq:predset}) 
\begin{equation*}
    P(\widehat{C}_n(\Xnew) = \emptyset |Y_i \in \{1,\dots, L\}) \leq \alpha
\end{equation*}
as $n\to \infty$ for the predictive set defined in (\ref{eq:predset}).
\end{corollary}

Corollary \ref{col:outlier} can be proved in the same way as Theorem \ref{thm:coverage}. It states that an inlier will be mistakenly classified as an outlier of the probability at most $\alpha$. 

\section{Implementation}
We construct $I_\ell$ and $G_\ell$ based on deep neural network architectures including Visual Geometry Group (VGG) \citep{simonyan2014very} and deep residual networks (Resnet) \citep{he2016deep}. Particularly, we parallelly apply the stochastic gradient algorithm for subsets of data. Take $\ell = 1$ as an example. We train $I_1(\cdot)$ to produce $\bZ_{i,1}$'s that follow multivariate Gaussian distribution for $X_i$'s from the 1st group, yielding the Chi-squared distributed $T_i^1$'s. To further reduce the size of the predictive set, we aim to make the presentation $\bZ_{i, 1}$ to be distinguishable to $\bZ_{i, \ell'}$ for $\ell'\in\{1,\dots,L\}\setminus \{1\}$.  We fine turn $I_\ell$ with respect to 
\begin{equation}\label{eq:Lpred}
\begin{split}
    L^{\rm pred}_\ell = & \bbE_{X\sim F_\ell}[ \log P(Y = \ell|I_\ell(X)) ] \\
    \quad & + \bbE_{X\nsim F_\ell}[\log(1-P(Y = \ell|I_\ell(X)))]
\end{split}
\end{equation}
where we use sample average of $\{X_i: Y_i\ne \ell\}$ to replace the expectation in (\ref{eq:Lpred}). Practically, we only need to randomly sample from $\{X_i: Y_i\ne \ell\}$ with the same size as $|\{X_i: Y_i\ne \ell\}|$.
We summarize the implementation into two Algorithms: Algorithm 1 is designed to learn the $I_\ell$ and $G_\ell$ for $\ell=1,\dots, L$ to calculate the p-values; Algorithm 2 is for constructing the predictive set. The computational cost is at most twice the computational cost for the classification model since the sample size is smaller for each class. 

\begin{algorithm}[ht]
\textbf{Input:} Test size $\alpha$, training dataset: $\{X_i\}_{i=1}^n$, training dataset labels: $\{Y_i\}_{i=1}^n$, test dataset: $\{X_i\}_{i=n+1}^{n+m}$, number of epochs $N$. 

\begin{algorithmic}[1]
\FOR{$\ell = 1, \dots, L$}
\STATE Initialize $G_\ell, D_{\ell}$ and $I_\ell$.
\FOR{epoch = $1, \dots, N$} 
\STATE Train $G_\ell, D_{\ell}$ and $I_\ell$ simultaneously on $\{X_i: Y_i = \ell, i = 1, \dots, n\}$ to minimize $V_{\rm total} = V_{G} + V_I + V_{\rm cycle}$.
\STATE Fine-tune $I_\ell$ on $\{X_i: Y_i \neq \ell, i = 1, \dots, n\}$ to minimize $L^{\rm pred}_\ell$ in (\ref{eq:Lpred}).
\ENDFOR
\STATE Apply $T^\ell(X)$ to $\{X_i: Y_i = \ell, i = 1, \dots, n\}$ to obtain $A_\ell$ in (\ref{eq:pool}).

\ENDFOR
\FOR{$i = n + 1, \dots, n+m$}
\STATE Based on non-conformity score pool $A_\ell$, we compute $p$-values $\bpi_i = \{ \pi_i^1, \pi_i^2, \dots, \pi_i^L \}$.
\ENDFOR
\end{algorithmic}
\textbf{Output:} P-values of all test observations: $\{\bpi_i\}_{i=n+1}^{n+m}$.
\caption{Flow-based Conformal Inference}\label{alg:fci}
\end{algorithm}

\begin{algorithm}[ht]
\textbf{Input:} Test size $\alpha$, $\{\bpi_i\}_{i=n+1}^{n+m}$ from Algorithm \ref{alg:fci}.

\begin{algorithmic}[1]
\FOR{$i = n+1, \dots, n+m$}
\STATE Construct predictive set for $i$th sample $\widehat{C}_n(X_i) = \{\ell: \pi_i^\ell \geq \alpha \}$.
\ENDFOR
\end{algorithmic}
Conduct statistical inference based on $\{\widehat{C}_n(X_i)\}_{i=n+1}^{n+m}$.
\caption{\FCI for predictive set}\label{alg:fcips}
\end{algorithm}

\section{Numerical Results}

\begin{table*}[ht]
\centering
\begin{tabular}{@{}cccccccc@{}}
\toprule
\multirow{2}{*}{Contamination rate} & \multirow{2}{*}{Model} & \multicolumn{3}{c}{Coverage}                        & \multicolumn{3}{c}{Size Error}             \\ \cmidrule(l){3-8} 
                                    &                        & Scaling         & APS             & FCI             & Scaling         & APS    & FCI             \\ \midrule
\multirow{3}{*}{0\%}                & VGG16                  & 0.9281          & 0.9458          & \textbf{0.9489} & \textbf{0.1466} & 0.3709 & 0.1523          \\
                                    & ResNet18               & \textbf{0.9488} & 0.9468          & 0.9454          & 0.2396          & 0.3436 & \textbf{0.0457} \\
                                    & ResNet34               & 0.8753          & \textbf{0.9460} & 0.9458          & 0.0740          & 0.3469 & \textbf{0.0566} \\ \midrule
\multirow{3}{*}{5\%}                & VGG16                  & 0.8292          & 0.8975          & \textbf{0.9424} & \textbf{0.1051} & 0.4307 & 0.1708          \\
                                    & ResNet18               & 0.8430          & 0.8989          & \textbf{0.9507} & 0.1451          & 0.3834 & \textbf{0.0520} \\
                                    & ResNet34               & 0.8281          & 0.8982          & \textbf{0.9473} & 0.0877          & 0.3681 & \textbf{0.0558} \\ \midrule
\multirow{3}{*}{10\%}               & VGG16                  & 0.7872          & 0.8506          & \textbf{0.9412} & 0.2208          & 0.5899 & \textbf{0.1288} \\
                                    & ResNet18               & 0.7971          & 0.8515          & \textbf{0.9489} & 0.3351          & 0.6995 & \textbf{0.0511} \\
                                    & ResNet34               & 0.7831          & 0.8500          & \textbf{0.9497} & 0.2277          & 0.6166 & \textbf{0.0504} \\ \bottomrule
\end{tabular}
\caption{\textbf{Results on Fashion-MNIST} We report the coverage and size error of optimal \Scaling, \APS, \FCI sets under different contamination rates. Top-1 and top-3 accuracies are also recorded (Same for \Scaling and \APS). Each entry is the mean of 50 independent trials. We use the bold font for coverages closest to $1-\alpha$ and the most minor size errors.}
\label{tab:classification}
\end{table*}

\begin{table*}[ht]
\centering
\begin{tabular}{@{}cccccccc@{}}
\toprule
\multirow{2}{*}{Contamination rate} & \multirow{2}{*}{Model} & \multicolumn{3}{c}{Coverage}                & \multicolumn{3}{c}{Size Error}     \\ \cmidrule(l){3-8} 
                                    &                        & Scaling & APS             & FCI             & Scaling & APS    & FCI             \\ \midrule
\multirow{3}{*}{0\%}                & VGG16                  & 0.8846  & 0.9502          & \textbf{0.9500} & 3.5131  & 4.3924 & \textbf{3.1034} \\
                                    & ResNet18               & 0.7434  & \textbf{0.9501} & 0.9301          & 1.0563  & 4.0446 & \textbf{0.0863} \\
                                    & ResNet34               & 0.6311  & \textbf{0.9499} & 0.9411          & 0.5021  & 4.6576 & \textbf{0.0591} \\ \midrule
\multirow{3}{*}{5\%}                & VGG16                  & 0.7094  & 0.9017          & \textbf{0.9123} & 0.8671  & 2.9450 & \textbf{0.1134} \\
                                    & ResNet18               & 0.6612  & 0.9020          & \textbf{0.9238} & 0.7589  & 4.1094 & \textbf{0.0769} \\
                                    & ResNet34               & 0.6730  & 0.9013          & \textbf{0.9244} & 0.2990  & 2.6541 & \textbf{0.0755} \\ \midrule
\multirow{3}{*}{10\%}               & VGG16                  & 0.6659  & 0.8555          & \textbf{0.9117} & 0.7539  & 2.9526 & \textbf{0.1094} \\
                                    & ResNet18               & 0.6249  & 0.8532          & \textbf{0.9174} & 0.8909  & 4.4349 & \textbf{0.0938} \\
                                    & ResNet34               & 0.6396  & 0.8559          & \textbf{0.9095} & 0.4317  & 2.9268 & \textbf{0.1015} \\ \bottomrule
\end{tabular}
\caption{\textbf{Results on CIFAR-10} These results are recorded similarly as in Table~\ref{tab:classification} except that each entry is the mean of 30 independent trials. We use the bold font for coverages closest to $1-\alpha$ and the most minor size errors.}
\label{tab:classification2}
\end{table*}

\begin{figure*}[ht]
    \centering
    \includegraphics[width = 0.7\textwidth]{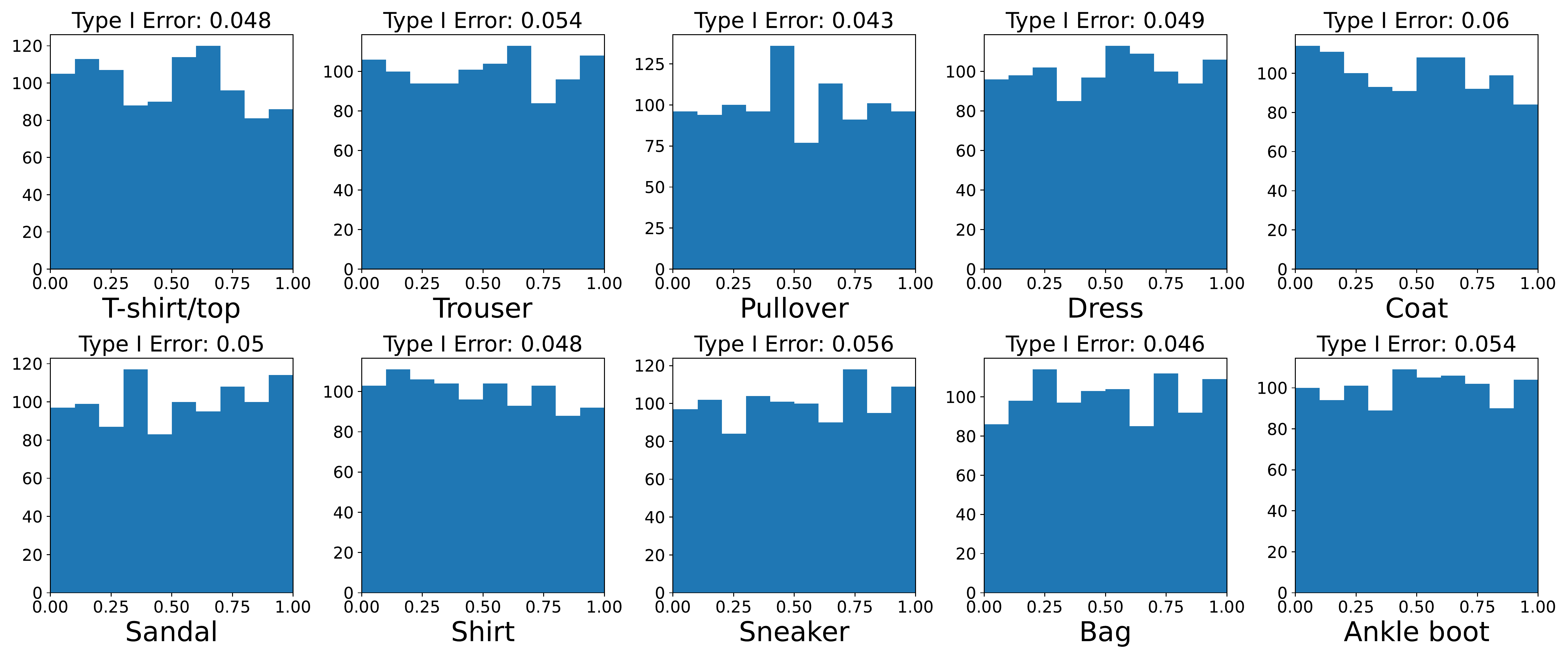}
    \caption{The histograms of $\{\pi^\ell_i: Y_i = \ell)\}$ for $\ell = 1,\dots,10$ on Fashion-MNIST. They are supposed to follow uniform distributions.}
    \label{fig:size}
\end{figure*}

In this section, we compare the performance of our proposed methods with existing works using various deep neural network architectures. 
We first compare with the method in \cite{platt1999} and \cite{guo2017calibration}, which constructs a predictive set by including classes from highest to lowest probability until their sum exceeds $1-\alpha$. 
A brief illustration of the method would be: if we have $s_1, s_2, \dots, s_L$ as the softmax output of a CNN, we order them first, producing ranked outputs $s_{(L)}, s_{(L-1)}, \dots, s_{(1)}$ and the corresponding indexes $i_1, i_2, \dots, i_L$. The predictive set $\widehat{C}_n(X_i)$ is $\{i_1, i_2, \dots, i_c\}$ if $\sum_{\ell = 1}^{c-1} s_{(L-\ell+1)} < \alpha$ and $\sum_{\ell = 1}^{c} s_{(L-\ell+1)} \geq \alpha$. We call this strategy the \Scaling method. Since the probabilities output by CNNs is known to be incorrect \citep{nixon2019measuring}, the coverage level usually can not be achieved.
We also compare our approach with the data-splitting conformal predictive method in \citet{Romano2020}, as Adaptive Prediction Sets (\APS). 
Their work can be applied to any black-box algorithm with a split-conformal calibration with a statistical guarantee to reach the coverage level under the exchangeable assumption.

In our numerical experiments, we first compare different methods under regular settings without outliers. Afterward, we test our model performance under different contamination rates, i.e., with outliers in varying percentages. Our experiments are based on two benchmark datasets: Fashion-MNIST (\url{https://github.com/zalandoresearch/fashion-mnist}) provided by \citet{xiao2017fmnist} and CIFAR-10 (\url{https://www.cs.toronto.edu/~kriz/cifar.html}) provided by \citet{Krizhevsky09learning}. For Fashion-MNIST, there are ten classes, each of which has 6000 points in the training dataset and 1000 points in the test dataset. For CIFAR-10, there are also ten classes. However, each class has only 5000 training observations and 1000 test observations. 

\subsection{Constructing Predictive Sets on Fashion-MNIST and CIFAR-10}
{\bf Experiment settings} We consider two popular neural network architectures (VGG16, ResNet18, ResNet34) to implement \Scaling, \APS, and our \FCI approach.
We use Adam optimizer \citep{kingma2014adam} in all experiment settings. Note that we test the performance of three methods when exposed to outliers. In order to reflect the degree of contamination of the dataset, we define the contamination rate as the ratio of the outlier sample size to the total sample size in the test set. We conduct experiments with 0\%, 5\%, and 10\% contamination rates. Under the 0\% contamination rate, there are no outliers; thus, the training set and test set have the same classes (from 0 to 9). However, under 5\% or 10\% contamination rate, one of the ten classes is removed in training sets but used as outliers in the test sets. Specifically, we removed {\itshape Ankle boot} in Fashion-MNIST and {\itshape Horse} in CIFAR-10 from training data. Since these two classes are not used for training, we treat them as outliers in the testing phase. We set the predefined coverage level $1- \alpha = 0.95$ in all settings. 
We compared the performance of three methods mainly on two criteria:
\begin{equation*}
\begin{split}
    \mbox{coverage} & = \frac{1}{m}\sum_{i=n+1}^{n + m} \left[\mathbbm{1}_{\{Y_i \in \overline{\cO} \}} \cdot \mathbbm{1}_{\{Y_i \in \widehat{C}_n(X_i)\}} \right.\\
    & \qquad \left.+ \mathbbm{1}_{\{Y_i \in \{\cO\}\}} \cdot \mathbbm{1}_{\{\widehat{C}_n(X_i) = \emptyset\}}\right].
\end{split}
\end{equation*}
\begin{equation*}
    \mbox{size error} =  \frac{1}{m} \sum_{i=n+1}^{n+m} \left[|\widehat{C}_n(X_i)| - \mathbbm{1}_{\{Y_i \in \{\cO\}\}} \right].
\end{equation*}
where $\mathbbm{1}_{\{\cdot\}}$ is an indicator function, $\cO$ denotes the set of class labels that have not been seen in the training data, and $\overline{\cO}$ is its complement. 

We first analyze the performance of all methods on the Fashion-MNIST dataset. As shown in the first three rows of table \ref{tab:classification}, \Scaling shows lower coverage probability under VGG16 and ResNet34. \APS achieves about 95\% coverage probability in all cases. The cost of achieving high coverage is the large size error. Compared to \Scaling or \APS,  \FCI also covers the actual class labels at around 95\% times. Moreover, \FCI achieves the smallest size error under ResNet18 and ResNet34 and achieves the second-best size error under VGG16. Next, we analyze the performance of methods when exposed to contamination. The coverages of both \Scaling and \APS drop from 0.95 to around 0.9 as the contamination rate increases to 5\%. When the contamination rate increases to 10\%, the coverages decline to around 85\%. Meanwhile, the average size error increases correspondingly. However, The performance of \FCI is not affected much by the outliers. It achieves the highest coverage and the smallest size error under all three models.  

Then we use CIFAR-10 dataset as another example to examine the performance of different methods. Similarly, \Scaling can not achieve the coverage probability. However, \APS still has coverages very close to the pre-specified level on all three models without contamination. The coverage of \FCI on VGG16 is desirable and slightly lower than 95\% under two ResNet architectures. 
Under the contaminated cases, the coverage drops significantly for the \Scaling method. 
\APS's coverage drops around 5\% and 10\%. \FCI has the least drop of coverage in all settings and maintains at least  91\% coverage when the contamination is 10\%.
For contaminated settings, the size errors of \FCI are around 40\%, 70\%, and 110\% smaller than the size errors of \APS under three different network architectures correspondingly. In sum,  \FCI shows more robust performance compared to either \Scaling or \APS on CIFAR-10 dataset.

In addition, we numerically verify the distribution of our proposed $p$-value is uniform in $(0,1)$. Figure \ref{fig:size} shows the histograms of $\pi_i^\ell$ under the null hypothesis, i.e., $\{\pi^\ell_i: Y_i = \ell \}$. For all classes, we see the $p$-values distribute uniformly, and the type-I error rates are around 5\%. It demonstrates the \FCI's ability to measure the randomness in the sample space $\cX$ and avoid the problem of overconfidence or underconfidence in classical CNN output \citep{nixon2019measuring}. 

\section{Discussion}

In modern applications, the testing data has a distributional shift from training data or contains contamination that violates the exchangeable assumption for data-splitting conformal prediction. Our approach quantifies the uncertainty through the conditional adversarial flow, which is robust to the distributional change and outliers. Since we need to train a flow model for each class, the overall architecture is more complex than the classical classification model. In future work, we aim to improve the network architecture, which combines all the mappings into a single process.  We expect to extend our work to a wider variety of real-world problems with a large number of classes. In addition, our proposed theory on building a valid $p$-value can provide a general framework for multiple inference tasks such as the two-sample test and goodness-of-fit test.

\bibliography{ref}
\bibliographystyle{plainnat}

\end{document}